\documentclass[lettersize,journal]{IEEEtran}
\usepackage{amsmath,amsfonts}
\usepackage{algorithmic}
\usepackage{algorithm}
\usepackage{array}
\usepackage[caption=false,font=normalsize,labelfont=sf,textfont=sf]{subfig}
\usepackage{textcomp}
\usepackage{stfloats}
\usepackage{url}
\usepackage{verbatim}
\usepackage{graphicx}
\usepackage{cite}
\usepackage{multirow}
\usepackage{booktabs}
\usepackage{amssymb}
\usepackage[table]{xcolor}  
\usepackage[most]{tcolorbox}
\usepackage{enumitem}
\usepackage{framed}
\usepackage{amsmath}
\usepackage[colorlinks,
            linkcolor=red,
            anchorcolor=blue,
            citecolor=green,
            urlcolor=black
            ]{hyperref}
\usepackage[utf8]{inputenc}
\usepackage{pifont}
\usepackage{makecell}
\usepackage{hyperref}
\newcommand{\email}[1]{\href{mailto:#1}{#1}}

\begin{document}

\title{PS-ReID: Advancing Person Re-Identification and Precise Segmentation with Multimodal Retrieval}

 

\author{Jincheng Yan, Xiaoyan Luo,~\IEEEmembership{Member,~IEEE}, Yun Wang, and Yu-Wing Tai,~\IEEEmembership{Senior Member,~IEEE}
\thanks{Jincheng Yan, Xiaoyan Luo and Yun Wang are with School of Astronautics, Beihang University, Beijing, China (e-mail: \email{yan\_jc@buaa.edu.cn}; \email{luoxy@buaa.edu.cn}; \email{wangyunbuaa@buaa.edu.cn})}
\thanks{Yu-Wing Tai is with the Department of Computer Science, Dartmouth College, Hanover, NH, USA. (e-mail: \email{yuwing@gmail.com})}
\thanks{\textit{Corresponding author: Xiaoyan Luo}}
}




\maketitle

\begin{abstract}

Person re-identification (ReID) plays a critical role in applications such as security surveillance and criminal investigations. Most traditional image-based ReID methods face challenges including occlusions and lighting changes, while text provides complementary information to mitigate these issues. However, the integration of both image and text modalities remains underexplored. To address this gap, we propose {\bf PS-ReID}, a multimodal model that combines image and text inputs to enhance ReID performance. In contrast to existing ReID methods limited by cropped pedestrian images, our PS-ReID focuses on full-scene settings and introduces a multimodal ReID task that incorporates segmentation, enabling precise feature extraction of the queried individual, even under challenging conditions such as occlusion. 
To this end, our model adopts a dual-path asymmetric encoding scheme that explicitly separates query and target roles: the query branch captures identity-discriminative cues, while the target branch performs holistic scene reasoning. Additionally, a token-level ReID loss supervises identity-aware tokens, coupling retrieval and segmentation to yield masks that are both spatially precise and identity-consistent. 
To facilitate systematic evaluation, we construct M\textsuperscript{2}ReID, currently the largest full-scene multimodal ReID dataset, with over 200K images and 4,894 identities, featuring multimodal queries and high-quality segmentation masks. Experimental results demonstrate that PS-ReID significantly outperforms unimodal query-based models in both ReID and segmentation tasks. The model excels in challenging real-world scenarios such as occlusion, low lighting, and background clutter, offering a robust and flexible solution for person retrieval and segmentation. All code, models, and datasets will be publicly available.

\end{abstract}


\section{Introduction}
\label{sec:intro}

\noindent
\IEEEPARstart{P}{erson} re-identification (ReID) \cite{ahmed2015improved, wang2019beyond, ye2021deep} aims to match individuals across large image galleries from non-overlapping cameras, which is essential for applications like security and criminal investigations. Traditional image-based ReID methods \cite{chen2017beyond, zou2023discrepant, zhu2024seas} suffer from occlusions \cite{hou2019vrstc, wang2022pose}, lighting variations \cite{lu2023illumination}, and pose changes \cite{liu2023learning, sarfraz2018pose}, leading to suboptimal performance, particularly in real-world scenarios. Multimodal approaches that integrate both image and text inputs offer a promising solution. Text can provide supplementary context, such as details on clothing or accessories, which may not be captured in the image, improving matching accuracy, especially in challenging scenarios like low-quality images or ambiguous visual cues.

While image- and text-based ReID systems have advanced significantly \cite{qin2024noisy, shao2023unified, bai2023rasa}, multimodal integration remains limited. Existing methods often fuse image and text features shallowly, underutilizing their complementary strengths and relying heavily on unimodal inputs, which limits robustness in complex real-world scenarios. 
Furthermore, current ReID frameworks are typically restricted to cropped pedestrian images and focus solely on identity matching between pre-selected instances, ignoring the practical yet challenging setting of full-scene images with multiple individuals and cluttered backgrounds. 
In this setting, a ReID system must separate the queried individual from complex surroundings and extract precise features, which can be effectively accomplished through segmentation. Coupling retrieval with fine-grained segmentation enhances practicality and enables downstream tasks such as trajectory estimation, behavior analysis, and human–object interaction understanding.

\begin{figure}
\begin{center}
\includegraphics[width=1.0\linewidth]{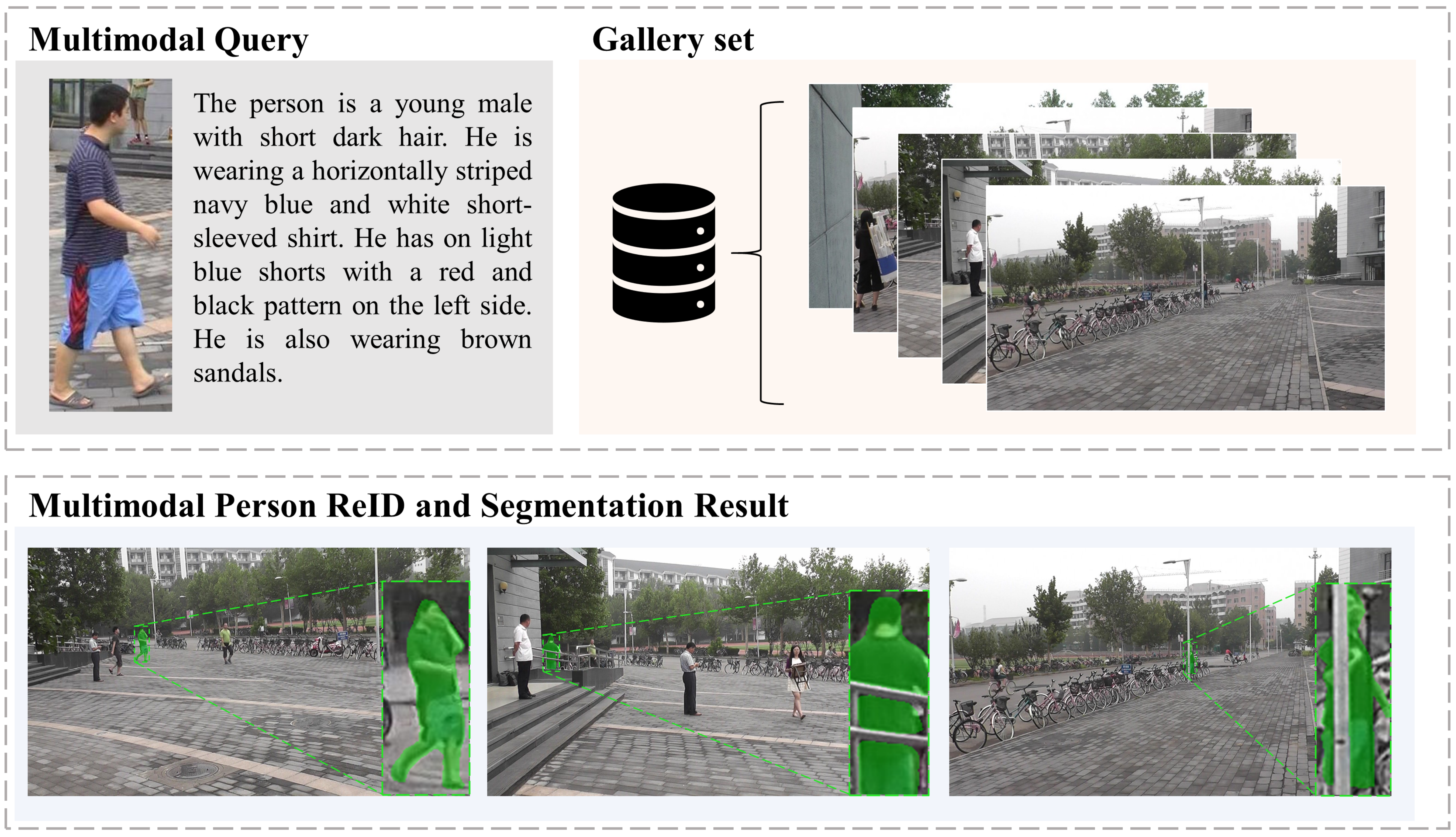}
\end{center}
\vspace{-0.15in}
   \caption{The goal of Multimodal Person ReID and Segmentation. Given multimodal query information, the model retrieves the matched individual from the gallery set and outputs its corresponding mask in the image.}
\label{fig:ProblemDefinition}\vspace{-0.15in}
\end{figure}

In this paper, we introduce a novel task, \textit{multimodal person ReID and segmentation}, as illustrated in Fig.~\ref{fig:ProblemDefinition}, which requires retrieving the correct individual from large, cluttered galleries, while simultaneously generating pixel-level masks under multimodal guidance. Unlike traditional ReID, this task addresses full-scene complexity with multiple people, occlusions, and background noise, demanding a unified framework that combines robust retrieval with fine-grained segmentation. To achieve this, the model must possess two key capabilities: 1) accurately retrieving gallery images containing the queried individual based on multimodal queries, and 2) generating identity-consistent segmentation masks within those images.

To this end, we propose {\bf PS-ReID} (Person-aware Segmentation and Re-identification Model), a large multimodal framework that unifies retrieval and segmentation within a single architecture. Departing from conventional approaches that process visual inputs in a uniform branch, PS-ReID introduces a dual-path asymmetric encoding strategy, where query and target images are aligned through independent adapters. This explicit role separation avoids semantic entanglement and enables the model to extract identity-discriminative cues from the query while conducting holistic scene-level reasoning on the target. Compared to conventional single-branch designs, this mechanism substantially improves robustness under full-scene complexity, including occlusion, background clutter, and varying illumination. 

Beyond architectural design, PS-ReID introduces a token-based ReID loss that directly supervises identity-aware tokens generated by the LMM. In contrast to conventional embedding-level objectives, this token-level supervision aligns identity representation learning with the reasoning space of the model, yielding embeddings that are both highly discriminative for retrieval and identity-consistent for downstream mask generation.

Additionally, PS-ReID employs structured prompting and unified supervision to bridge reasoning and localization. By transforming the LMM’s control tokens into semantic prompts for the segmentation module, high-level decisions are converted into pixel-level guidance, enabling precise and consistent mask generation even in challenging real-world scenarios. Collectively, these innovations establish PS-ReID as a novel and robust solution that integrates multimodal reasoning, retrieval, and fine-grained localization within a single framework.

To validate the effectiveness of PS-ReID, we introduce \textbf{M\textsuperscript{2}ReID} (Multimodal and Multi-task Person Re-identification Dataset), a new large-scale benchmark specifically designed for multimodal retrieval and segmentation. Unlike prior datasets focused on cropped pedestrian images, M\textsuperscript{2}ReID emphasizes full-scene complexity. It contains over 200K images and 4,894 identities, each paired with high-quality segmentation masks and descriptive text queries generated by SAM-2\cite{ravi2024sam}, and GPT-4o, and subsequently refined through human annotation. This dataset introduces realistic challenges closer to real-world applications and provides a solid foundation for advancing unified retrieval–segmentation research.

Our contributions are as follows:

\begin{itemize}
    \setlength{\itemsep}{0pt}
    \item We introduce the task of multimodal person re-identification and segmentation, which leverages both textual and visual queries to jointly tackle retrieval and fine-grained localization in full-scene conditions, beyond the cropped-image setting of conventional ReID.
    \item We present PS-ReID, a novel framework that tightly integrates retrieval and segmentation while supporting multimodal queries. It employs a dual-path asymmetric encoding to preserve the semantic roles of query and target inputs. Additionally, a token-level ReID loss is introduced to directly supervise identity-aware tokens, yielding embeddings that are both retrieval-discriminative and segmentation-consistent in complex scenes.
    \item We construct a large-scale multimodal pedestrian dataset, M\textsuperscript{2}ReID, featuring high-quality segmentation masks and descriptive textual annotations, which supports systematic evaluation under challenging real-world scenarios.
    
\end{itemize}

\section{Related Work}

\subsection{Person ReID}

Person ReID has been extensively studied under two main paradigms based on query modality: image-based ReID and text-based ReID.

\vspace{2mm}
\subsubsection{Image-based ReID}
This paradigm focuses on identifying individuals based solely on visual appearance. Early approaches mainly relied on handcrafted features and metric learning. 
With the advent of deep learning, CNN-based techniques have been widely adopted in person ReID to learn discriminative representations and optimize distance metrics. One of the earliest deep ReID models \cite{yi2014deep} applied a Siamese network with overlapping horizontal partitions and cosine similarity to learn discriminative embeddings for person matching. To further improve generalization, quadruplet loss was later introduced in \cite{chen2017beyond} to simultaneously reduce intra-class variation and increase inter-class margins. Circle Loss \cite{sun2020circle} unified classification and metric learning by enabling fine-grained optimization of pairwise similarities. More recent works, such as AdaSP \cite{zhou2023adaptive} and CA-Jaccard \cite{chen2024jaccard}, further improve ReID accuracy by adaptively selecting sparse positive pairs during training and enhancing Jaccard distance reliability with camera-aware neighbor modeling.

To address occlusion and misalignment, part-based methods extract local body features to enhance recognition robustness. PCB \cite{sun2018beyond} uniformly divides pedestrian images into horizontal stripes and learns independent features for each part, establishing a strong baseline with a simple yet effective design. SAN \cite{jin2020semantics} enhances part-level modeling by reconstructing semantic texture maps with an encoder–decoder alignment network. 
More recent efforts further mitigate occlusion by leveraging representative body part features \cite{somers2023body} and introducing shape-aligned supervision to enhance part-level discrimination \cite{zhu2024seas}.

Meanwhile, Transformer-based architectures have gained traction in person ReID due to their ability to model long-range dependencies and global context. TransReID \cite{he2021transreid} introduces jigsaw patch modules and side information embeddings to address spatial misalignment and camera variation. DCAL \cite{zhu2022dual} enhances fine-grained identity matching through global-local and pairwise cross-attention mechanisms. PHA \cite{zhang2023pha} further boosts ViT-based ReID by injecting high-frequency components via wavelet-based patch augmentation and contrastive learning.

\vspace{2mm}
\subsubsection{Text-based ReID} This paradigm extends person ReID by retrieving individuals from image galleries based on natural language descriptions \cite{li2017person}. 
Early text-based ReID methods primarily focused on enhancing modality-specific encoders, with the visual encoder evolving from VGG \cite{simonyan2014very} to ResNet \cite{he2016deep}, and the text encoder progressing from LSTM \cite{graves2012long} to BERT \cite{devlin2018bert}. Building upon these foundations, recent works leverage large-scale vision-language models such as CLIP \cite{radford2021learning, han2021text} and ALBEF \cite{li2021align, bai2023rasa} to significantly enhance cross-modal alignment in text-based person ReID.

Beyond backbone improvements, recent research has shifted toward finer-grained alignment strategies. While early methods aligned global embeddings of image and text in a shared space \cite{zheng2020dual, zhu2021dssl, shu2022see}, they often failed to capture discriminative local cues which are crucial for person ReID. To address this, later works focus on localized cross-modal interactions between visual regions and textual components. NAFS \cite{gao2021contextual} introduces a contextual non-local attention mechanism to align multi-scale visual and textual features, enabling more comprehensive cross-modal interactions. To further mitigate alignment ambiguity, LBUL \cite{wang2022look} incorporates distribution-aware manifold learning that jointly considers both visual and textual feature spaces, enhancing cross-modal consistency. Complementarily, PLOT \cite{park2025plot} shifts the focus to semantic part-level correspondence, employing slot attention for unsupervised body part discovery and text-guided dynamic attention to refine retrieval precision. These advances reflect a shift toward more localized and semantically grounded cross-modal alignment in text-based person ReID. 

Despite the progress in both image- and text-based ReID, most existing methods are still limited to unimodal queries and cropped person inputs, and the integration of both modalities remains underexplored. In contrast, the proposed PS-ReID supports multimodal joint queries and unifies retrieval with segmentation in full-scene images, substantially enhancing robustness and practicality in real-world scenarios.


\subsection{Large Multimodal Model}

Recent advances in large multimodal models (LMMs) have significantly improved the handling of multimodal tasks \cite{alayrac2022flamingo, ye2023mplug, zhu2023minigpt, liu2024visual, zhang2024llama}. The integration of large language models (LLMs) with multimodal inputs has become a rapidly growing area of research. Various strategies have been explored to align these inputs, including cross-attention modules \cite{alayrac2022flamingo}, Q-Former \cite{chen2024lion, dai2023instructblip, li2023blip}, and linear adapters \cite{cha2024honeybee, xuan2024pink}, to facilitate effective vision-language interaction. Many of these models are also equipped with task-specific heads to handle downstream visual tasks more effectively \cite{pi2023detgpt, rasheed2024glamm, zhang2024groundhog, kirillov2023segment}.

For instance, DetGPT \cite{pi2023detgpt} couples a fixed LMM with an open-vocabulary detector to support instruction-driven object detection, while GLaMM \cite{rasheed2024glamm} prompts a pixel encoder using LLM-generated descriptions to achieve pixel-level grounding. To enhance reasoning capabilities, recent models introduce learnable tokens \cite{lai2024lisa, xia2024gsva, ren2024pixellm, yan2024visa}. LISA \cite{lai2024lisa} integrates a segmentation token to inject segmentation capabilities into LMMs, while GSVA \cite{xia2024gsva} extends this by predicting rejection tokens to discard irrelevant targets. PixelLM \cite{ren2024pixellm} introduces a lightweight decoder and segmentation codebook for efficient multi-target mask generation. Additionally, VISA \cite{yan2024visa} extends the temporal dimension by introducing a text-guided frame sampler and an object tracker, thereby equipping LMM with video object segmentation and tracking capabilities. In contrast, our PS-ReID concentrates on enhancing LMM’s abilities in holistic scene reasoning and identity discrimination.

While these models exhibit impressive capabilities in general visual-related tasks, they are not specifically designed for fine-grained, person-centric applications such as re-identification and segmentation, particularly under multimodal query conditions. Although they can perform retrieval tasks based on user instructions, they typically support only text-based queries and fail to handle the image queries essential for ReID tasks. In contrast, PS-ReID enables joint visual-text queries, allowing accurate person retrieval and precise mask prediction, thus enhancing the practicality of multimodal ReID and segmentation. Furthermore, the introduction of the M\textsuperscript{2}ReID dataset addresses the lack of suitable datasets, providing a valuable foundation for multimodal ReID and segmentation research.

\section{Dataset}

\begin{table*}[ht]
\centering
\caption{The comparison of our M\textsuperscript{2}ReID with the prior datasets, where "N/A" indicates the corresponding
information are not available from the original dataset.}
\label{table:dataset_comparison}
\resizebox{2\columnwidth}{!}{ 
\begin{tabular}{c|c|c|c|c|c|c|c|c}

\specialrule{1.2pt}{0pt}{0pt}
    Dataset & Task & Year & Data Type & Query Modality & \# Images & \# Identities & \# Instances & Annotation \\
    \hline
    CUHK03\cite{li2014deepreid} & Image-based ReID & 2014 & Cropped & Image & 14,096 & 1,467 & 14,096 & ID \\
    Market1501\cite{zheng2015scalable} & Image-based ReID & 2015 & Cropped & Image & 32,668 & 1,501 & 32,668 & ID \\
    MSMT17\cite{wei2018person} & Image-based ReID & 2018 & Cropped & Image & 126K & 4,104 & 126K & ID \\
    LPW\cite{song2018region} & Image-based ReID & 2018 & Cropped & Image & 590K & 2,731 & 590K & ID \\
    SYSU30k\cite{wang2020weakly} & Image-based ReID & 2020 & Cropped & Image & 29M & 30K & 29M & Weak ID \\
    Luperson\cite{fu2020unsupervised} & Image-based ReID & 2021 & Cropped & Image & 4M & \textgreater200K & 4M & N/A \\
    Luperson-NL\cite{fu2022large} & Image-based ReID & 2022 & Cropped & Image & 10M & 433K & 10M & Noisy ID \\
    MEVID\cite{davila2023mevid} & Image-based ReID & 2023 & Cropped & Image & 10M & 158 & 1.7M & ID \\
   \hline
   \hline
    CUHK-PEDES\cite{li2017person} & Text-based ReID & 2017 & Cropped & Text & 40,206 & 13,003 & 40,206 & Text \\
    ICFG-PEDES\cite{ding2021semantically} & Text-based ReID & 2021 & Cropped & Text & 54,522 & 4,102 & 54,522 & Text \\
    RESPReid\cite{zhu2021dssl} & Text-based ReID & 2021 & Cropped & Text & 20,505 & 4,101 & 20,505 & Text \\
    Luperson-T\cite{shao2023unified} & Text-based ReID & 2023 & Cropped & Text & 1.3M & N/A & 1.3M & Text \\
   \hline
   \hline
   CUHK-SYSU\cite{xiao2017joint} & Person Search & 2017 & Full-scene & Image & 18,184 & 8,432 & 99,809 & ID/Bbox \\
   PRW\cite{zheng2017person} & Person Search & 2017 & Full-scene & Image & 11,816 & 932 & 34,304 & ID/Bbox \\
   LSPS\cite{zhong2020robust-APNet} & Person Search & 2020 & Full-scene & Image & 51,836 & 4,067 & 60,433 & ID/Bbox \\
   MovieNet-PS\cite{qin2023movienet} & Person Search & 2023 & Full-scene & Image & 160K & 3,000 & 274K & ID/Bbox \\
   \hline
   \hline
   \rowcolor{gray!20}
   M\textsuperscript{2}ReID(ours) & \makecell[c]{Multimodal Person ReID\\and Segmentation} & 2025 & Full-scene & Image/Text & 204K & 4,894 & 313K & ID/Text/Bbox/Mask \\
   
\specialrule{1.2pt}{0pt}{0pt}

\end{tabular}
}
\end{table*}

Although numerous datasets have been proposed for person ReID, as shown in Table \ref{table:dataset_comparison}, they still suffer from significant limitations:

\begin{itemize}
    \setlength{\itemsep}{0pt}
    \item Most existing datasets~\cite{li2014deepreid,zheng2015scalable,wei2018person,song2018region,wang2020weakly} only provide cropped pedestrian images, without the original scene context. This substantially hinders the modeling of contextual semantics, thereby limiting the development of higher-level tasks such as instance segmentation and multimodal retrieval.
    \item Some datasets~\cite{xiao2017joint,zheng2017person,zhong2020robust-APNet,qin2023movienet} include full-scene images, but they are either limited in scale or suffer from insufficient annotations, which restricts their applicability in comprehensive multimodal person ReID and segmentation tasks.
\end{itemize}

To address the above limitations, we construct a real-world dataset tailored for multimodal person ReID and segmentation. We design a data annotation pipeline based on publicly available video data filmed in diverse real-world scenarios. The dataset is constructed through a rigorous selection process and further refined with manual calibration to ensure annotation quality. The dataset offers substantial improvements in both scale and annotation richness over existing benchmarks.

\subsection{Data annotation pipeline}

The annotation pipeline of M\textsuperscript{2}ReID dataset can be summarized in four steps. First, we obtain the raw video data from Luperson-NL\cite{fu2022large} and MEVID\cite{davila2023mevid}, and apply the BoostTrack\cite{stanojevic2024boosttrack} algorithm for multi-object tracking to preliminarily filter and collect raw images. Subsequently, we employ 10 experienced annotators to perform quality control and filter the collected images, followed by the annotation of pedestrian identities and bounding boxes. Afterwards, we utilize SAM-2\cite{ravi2024sam} and GPT-4o to obtain pedestrian segmentation masks and corresponding textual descriptions. Finally, we further refine the dataset through manual inspection and correction of inaccurate annotations. 

\vspace{2mm}
\subsubsection{Data Collection}

To develop a comprehensive M\textsuperscript{2}ReID dataset, it is essential to include data captured from diverse real-world scenarios. For this purpose, we collect the original video data from Luperson-NL\cite{fu2022large} and MEVID\cite{davila2023mevid}, which span a wide range of environments, lighting conditions, and pedestrian appearances. From the raw videos, we first perform frame sampling to reduce redundancy. We then apply the BoostTrack\cite{stanojevic2024boosttrack} algorithm for multi-object tracking, and select qualified image samples based on the tracking results. Specifically, we discard detection results with confidence scores below 0.5, bounding box height under 75 pixels or width under 25 pixels. Finally, Non-Maximum Suppression (NMS)\cite{neubeck2006efficient} is applied to remove redundant detections and avoid trajectory duplication.

\vspace{2mm}
\subsubsection{Manual Annotation}

We engage 10 experienced annotators to manually filter and label the collected images. During this process, annotators thoroughly inspect each image based on tracking results, removing low-quality or ambiguous samples caused by occlusion, blur, or incorrect tracking. For qualified images, they annotate precise pedestrian bounding boxes along with corresponding identity IDs. To ensure high annotation accuracy and consistency across annotators, the annotation is conducted in batches: each annotator's work must pass a thorough quality check by a dedicated reviewer before gaining access to the next batch of images.

\vspace{2mm}
\subsubsection{Auto Labeling}

\begin{figure}
\begin{center}
\includegraphics[width=1.0\linewidth]{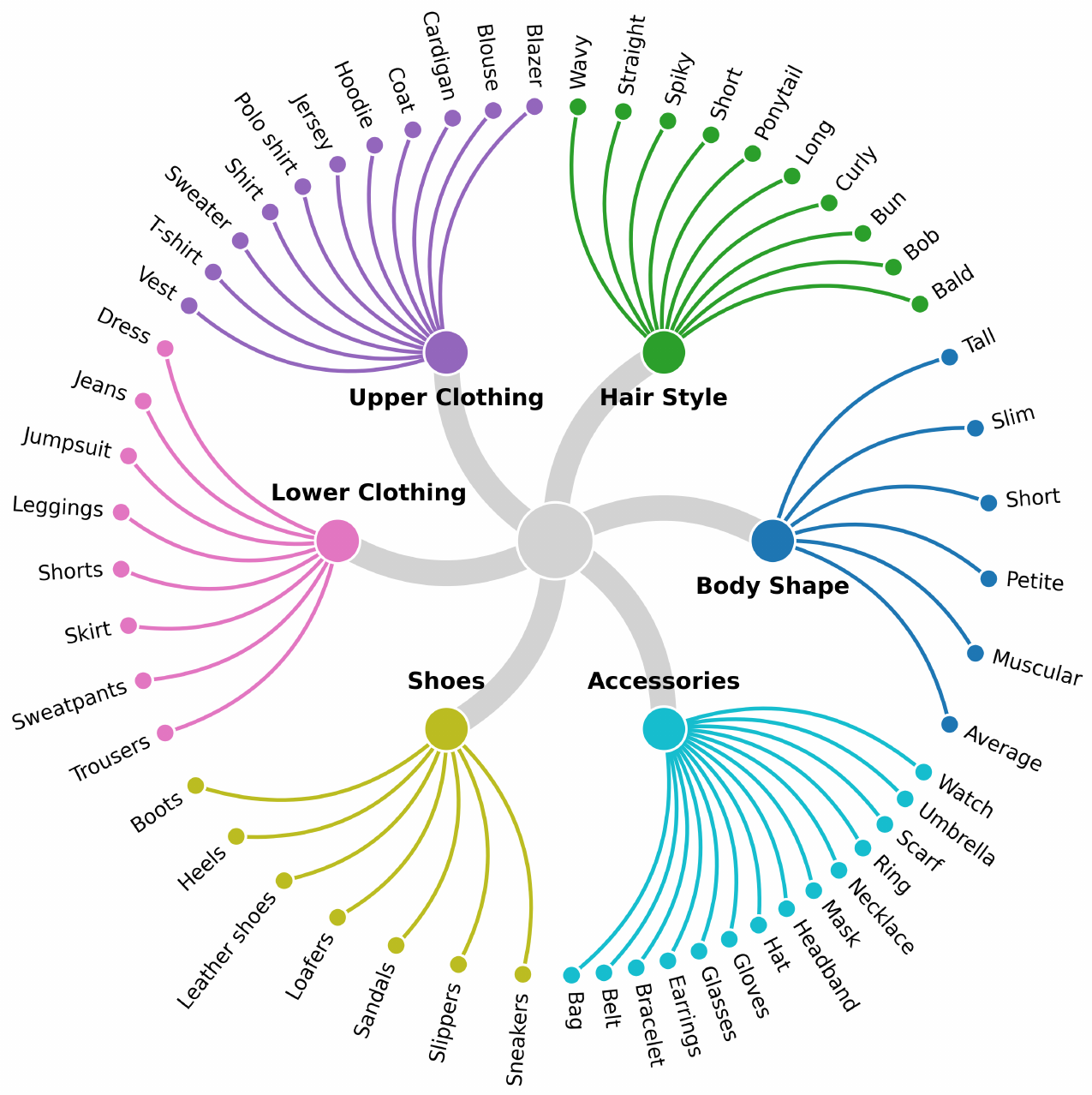}
\vspace{-0.35in}
\end{center}
   \caption{Visualization of person attributes in the M\textsuperscript{2}ReID dataset.}
\label{fig:AttributeTree}
\vspace{-0.15in}
\end{figure}

After manual annotation, we further enhance the dataset through automated labeling using advanced vision and language models. Specifically, the manually annotated pedestrian bounding boxes are used as spatial prompts to guide SAM-2 in automatically generating high-quality pixel-level segmentation masks. 
In parallel, we extract pedestrian image crops based on the bounding boxes and feed them into GPT-4o to generate corresponding textual descriptions. To ensure the accuracy and relevance of the generated captions, we first design an attribute taxonomy consisting of six categories, including hair style and body shape, as shown in Fig.~\ref{fig:AttributeTree}. The generation environment is set with a temperature of 0.5, and the following prompt is employed:

\vspace{-0.1em}
\begin{leftbar}
\noindent\small\textit{
``You are an expert in pedestrian appearance analysis. Based on the image below, please generate a concise and comprehensive textual description for a cross-modal person re-identification (Text-to-Image ReID) task. The description should cover all clearly visible attributes, including gender, body shape, hairstyle and hair color, upper and lower body clothing, shoes (color and style), and any visible accessories such as hats, glasses, masks, or bags. Only describe features that are clearly visible in the image—do not guess or infer any details that are occluded, partially shown, or out of frame. Avoid any subjective judgments or unrelated information such as pose, background, image quality, or camera angle. Write the description in a fluent paragraph, not as bullet points. Keep the language objective and neutral. The total length should be under 150 words, and only include the appearance description intended for model input.}

\vspace{0.3em}
\noindent\small\textit{- The pedestrian image to describe is: \texttt{<image>}.''}
\end{leftbar}
\vspace{-0.1em}

\vspace{2mm}
\subsubsection{Final Refinement}

Annotators carefully verify and refine the outputs from the automatic annotation stage. When SAM-2 fails to produce accurate segmentation masks for pedestrians, such as under occlusion or motion blur, annotators  manually correct the masks to align with the true pedestrian contours. Similarly, textual descriptions generated by GPT-4o are reviewed for hallucinations, factual inaccuracies, or inconsistencies with visual content. If issues are found, annotators revise the descriptions to ensure they are both semantically correct and contextually relevant. Through this process, we uphold a high-quality standard of our dataset, rendering M\textsuperscript{2}ReID a richly annotated and reliable benchmark for multimodal person analysis tasks.

\begin{figure*}
\begin{center}
\includegraphics[width=1.0\linewidth]{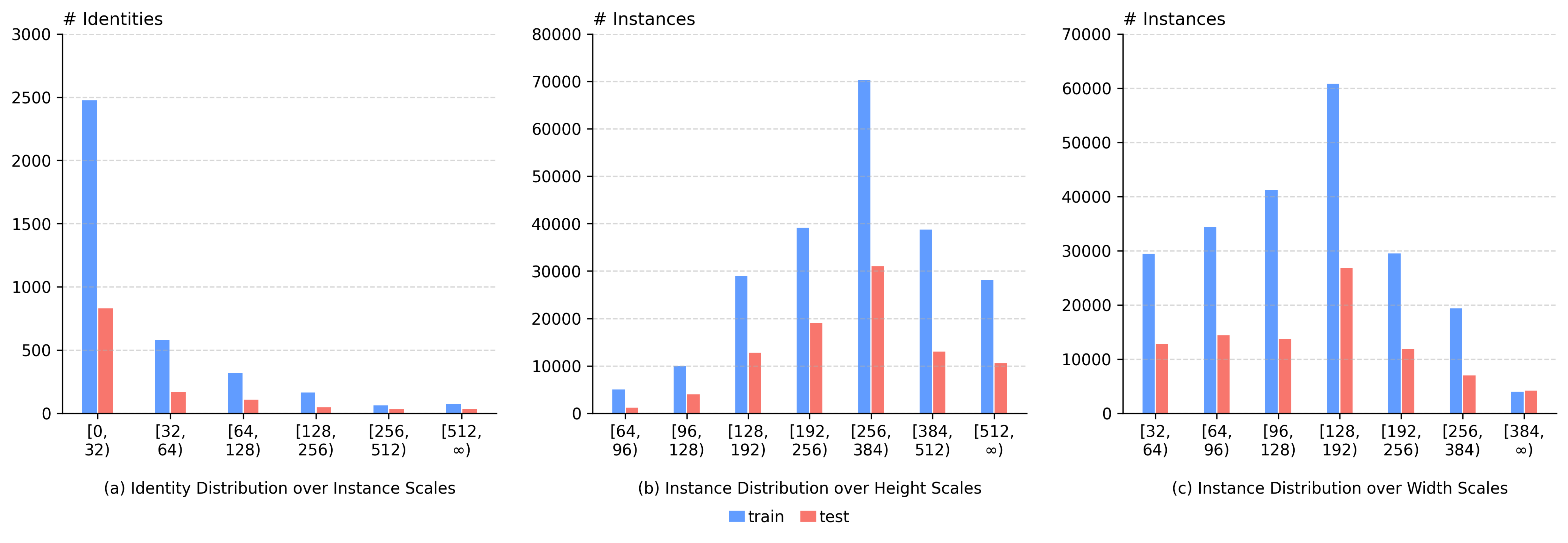}
\vspace{-0.35in}
\end{center}
   \caption{Data distribution of identities and target scales present in M\textsuperscript{2}ReID.}
\label{fig:DataDistribution}
\vspace{-0.15in}
\end{figure*}

\subsection{Dataset Characteristics}

\vspace{2mm}
\subsubsection{Relative Large-scale}

As demonstrated in Table \ref{table:dataset_comparison}, the proposed M\textsuperscript{2}ReID dataset demonstrates a significant advantage in scale compared to most existing datasets. Although datasets such as LPW\cite{song2018region} and SYSU30k\cite{wang2020weakly} contain a larger number of images, they are primarily designed for Re-ID tasks and only provide cropped pedestrian images without access to the original scenes. Additionally, datasets like Luperson provide original video frames, but their annotations are insufficient to support multimodal person re-identification and segmentation tasks. In contrast, M\textsuperscript{2}ReID contains 204,980 images, 313,159 instances, and 4,894 identities, with rich and multi-level annotations including identity labels, bounding boxes, segmentation masks, and textual descriptions. In terms of image quantity, instance volume and annotation completeness, M\textsuperscript{2}ReID stands out as one of the most comprehensive datasets currently available.

\vspace{2mm}
\subsubsection{Comprehensively Labeled}

The M\textsuperscript{2}ReID dataset is equipped with detailed annotations across multiple modalities. Each pedestrian instance is labeled with a unique identity ID, a bounding box, a segmentation mask, and a corresponding textual description. Fig.~\ref{fig:DataDistribution} presents the annotation statistics: the identity distribution exhibits a long-tailed pattern, and bounding-box sizes span diverse scales, from small distant individuals to large close-up subjects, reflecting the challenges of real-world surveillance.

\begin{figure}
\begin{center}
\includegraphics[width=0.8\linewidth]{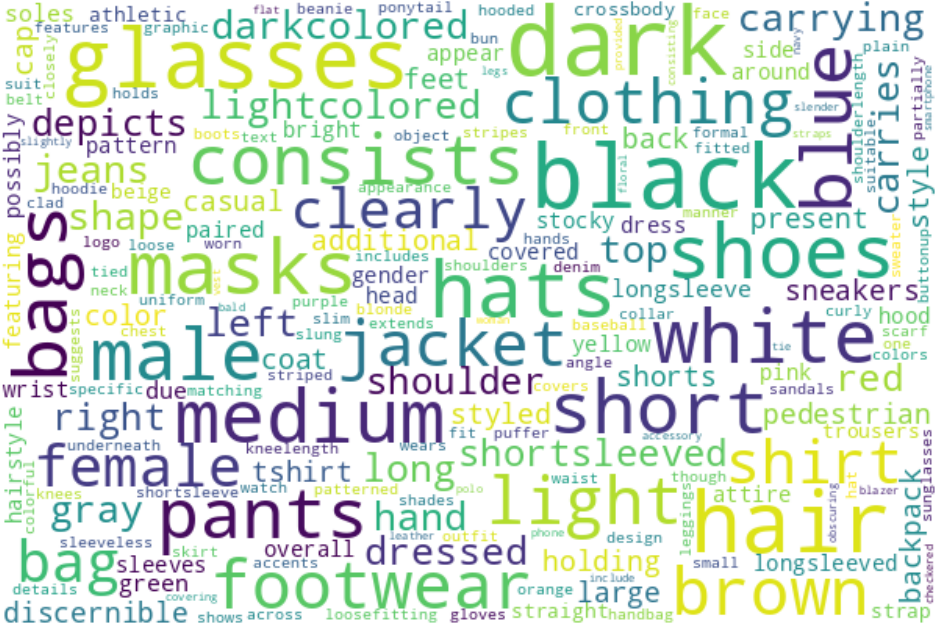}
\vspace{-0.15in}
\end{center}
   \caption{Word cloud visualization of textual annotations in the M\textsuperscript{2}ReID dataset.}
\label{fig:WordCloud}
\vspace{-0.15in}
\end{figure}

In addition to visual annotations, M\textsuperscript{2}ReID provides fine-grained textual descriptions for each identity. Based on the taxonomy shown in Fig.~\ref{fig:AttributeTree}, these descriptions are automatically generated with GPT-4o and refined through manual review. Fig.~\ref{fig:WordCloud} further analyzes the lexical composition of these descriptions, showing that high-frequency terms align well with the attribute schema. Overall, the textual annotations are semantically expressive, visually grounded, and consistent with the structured taxonomy, providing valuable multimodal cues for person ReID.

\vspace{2mm}
\subsubsection{Scene-diverse}

\begin{figure}
\begin{center}
\includegraphics[width=1.0\linewidth]{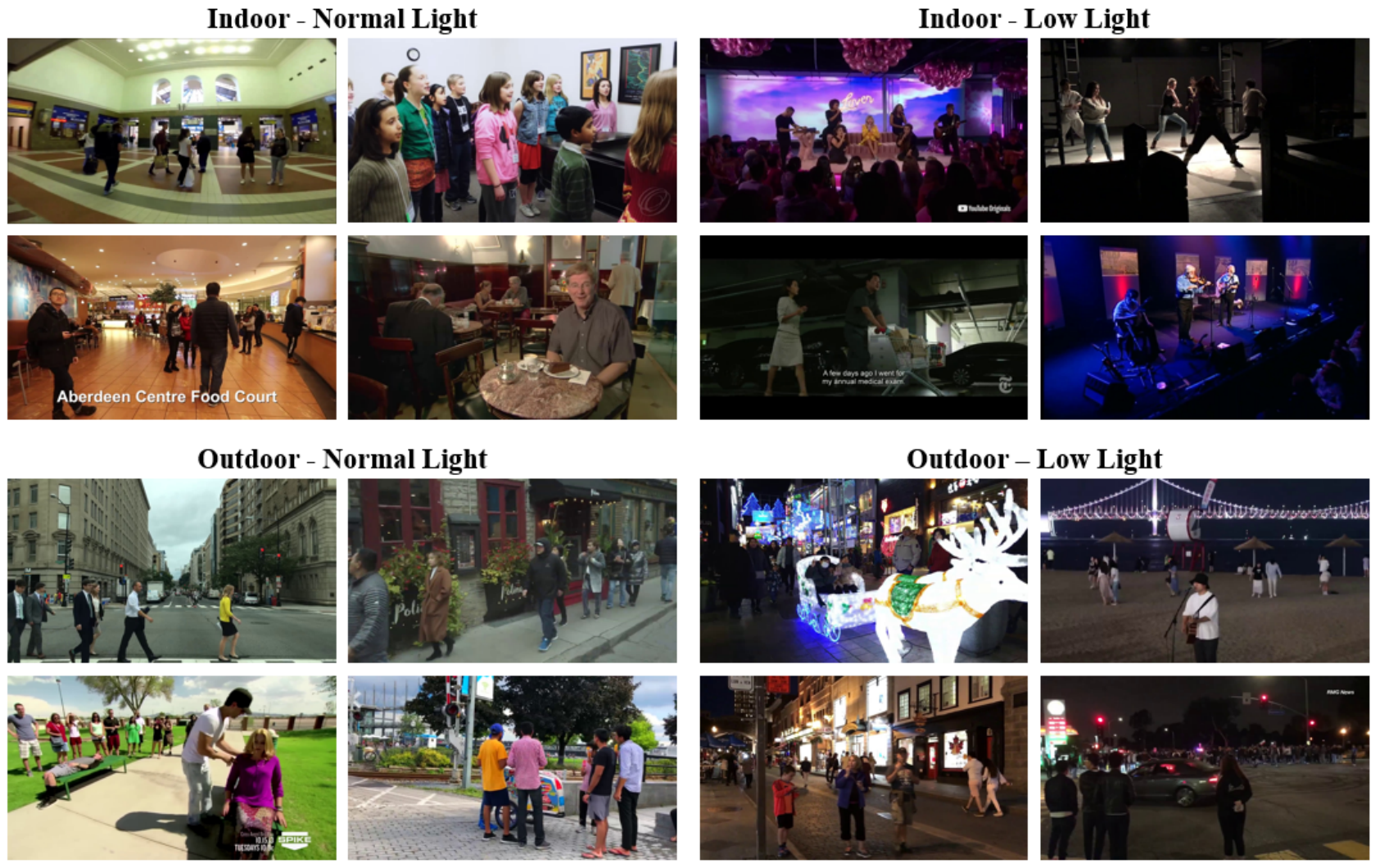}
\vspace{-0.35in}
\end{center}
   \caption{Examples of diverse lighting and scene conditions in the M\textsuperscript{2}ReID dataset.}
\label{fig:DataSceneDemo}
\vspace{-0.15in}
\end{figure}

The M\textsuperscript{2}ReID dataset encompasses a wide variety of scene conditions. Unlike existing datasets that are typically confined to well-lit outdoor environments or fixed camera setups, our dataset spans a broader range of real-world conditions. It includes both normal and low-light settings, as well as indoor and outdoor locations, offering a more realistic and challenging benchmark. Fig.~\ref{fig:DataSceneDemo} illustrates samples captured under diverse lighting and scene conditions, reflecting the dataset’s adaptability to complex environments and reinforcing its suitability for real-world applications.

\begin{figure*}
\begin{center}
\includegraphics[width=1.0\linewidth]{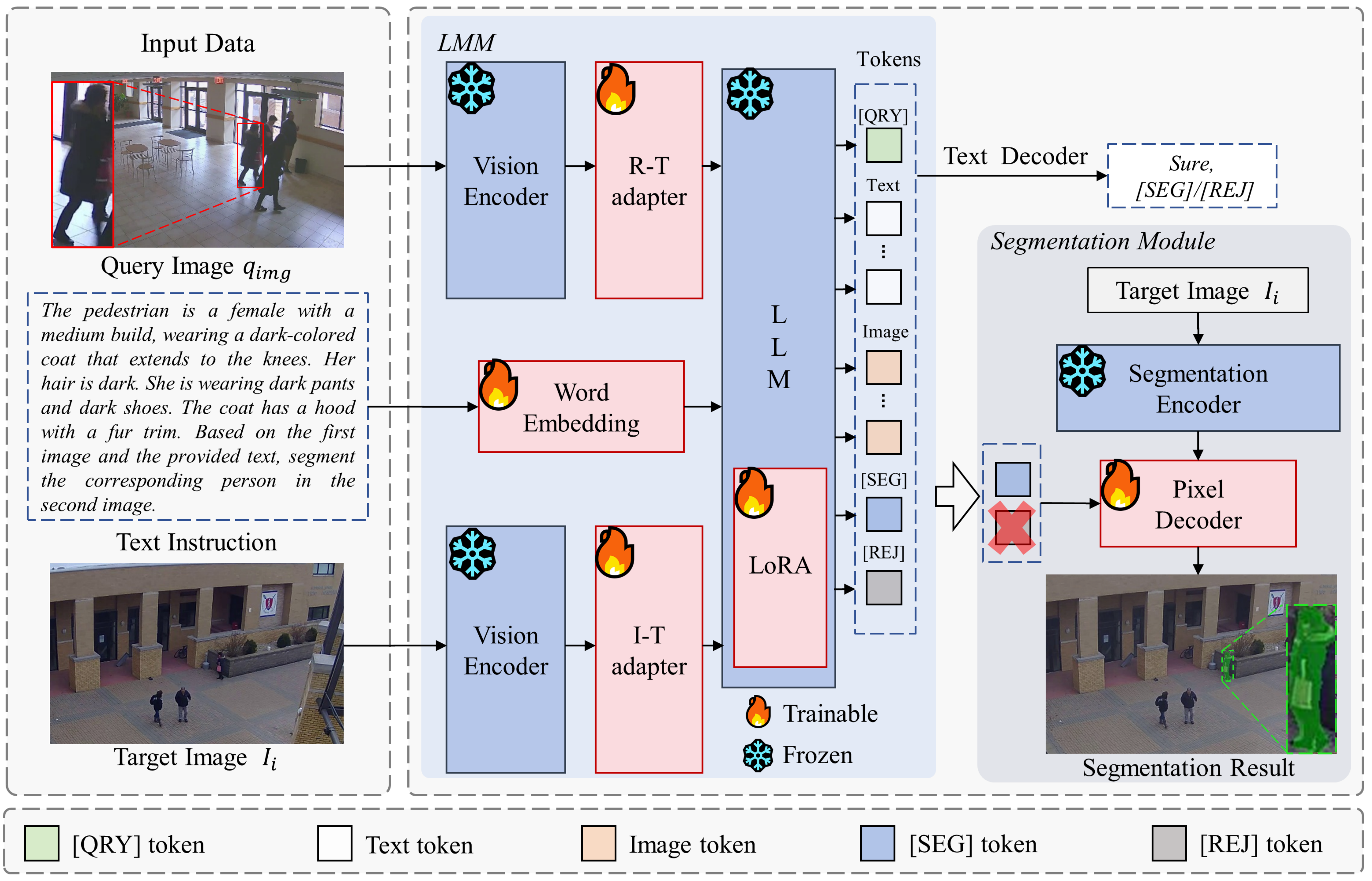}
\vspace{-0.25in}
\end{center}
   \caption{Overview of the proposed PS-ReID. The Large Multimodal Model (LMM) encodes both the query image and the target image, and concatenates the tokenized text tokens together. The Large Language Model (LLM) receives the concatenated embeddings and outputs either a [SEG] or [REJ] token to determine whether the queried individual is present in the target image. When a [SEG] token is generated, the segmentation module is activated, and the pixel decoder produces the final segmentation mask for the target individual.}
\label{fig:ModelArch}
\vspace{-0.15in}
\end{figure*}

\section{Multimodal Person ReID and Segmentation}

\subsection{Problem Definition}
The goal of Multimodal Person ReID and Segmentation is to retrieve the correct individual from the gallery that matches the query information and generate corresponding binary segmentation masks, as illustrated in Fig.~\ref{fig:ProblemDefinition}. Specifically, given a query image \( I^q \), a query text \( t^q \), and an image gallery \( \mathcal{G} = \{ I^t_i \}_{i=1}^{N} \), where \( N \) is the number of images in the gallery, our model \( \mathcal{F}\) determines whether the queried individual appears in each target image: 

\begin{equation}
l_i^q = \mathcal{F}\!\left(\left(I^q, t^q\right), I_i^t \right), \quad i=1,\dots,N
\end{equation}

\noindent
where \( l^q_i \in \{0, 1\} \) indicates whether the target image \( I^t_i\) contains the same individual described by the query. If \( l^q_i = 1 \), the model proceeds to generate a precise segmentation mask for the identified individual. The objective is to retrieve all images from the gallery that contain the queried individual and generate precise segmentation masks for them.

\subsection{PS-ReID}

We propose PS-ReID, a multimodal framework that unifies person retrieval and segmentation. As illustrated in Fig.~\ref{fig:ModelArch}, the architecture is composed of two core components: a large language model (LLM) that interprets the multimodal query inputs via an asymmetric dual-path design, and a segmentation module that produces accurate binary masks for the matched individual. 

The textual query is incorporated through the standard prompt-based strategy of LMMs, whereas our novelty lies in the asymmetric treatment of visual inputs: the query and target images are encoded via independent branches, thereby explicitly preserving their distinct semantic roles and overcoming the limitations of uniform single-branch designs. Guided by carefully crafted prompts, the LMM infers whether the target image contains the queried individual. Distinct from existing methods, PS-ReID further integrates multimodal query support with a tightly coupled retrieval–segmentation strategy, uniting the reasoning ability of the LLM with the spatial localization strength of the segmentation module to achieve robust and precise person retrieval with pixel-level segmentation across diverse conditions.

\vspace{2mm}
\subsubsection{Large Multimodal Model}

Unlike conventional single-branch designs that encode all visual inputs uniformly, the large multimodal model (LMM) in PS-ReID introduces a role-specific asymmetric alignment that explicitly preserves the distinct semantic roles of the query and target images, enabling joint interpretation of visual and textual cues. Specifically, two dedicated adapters are employed: a Region-to-Text adapter ($\phi_R$) for the query image and an Image-to-Text adapter ($\phi_I$) for the target image. This asymmetry allows the model to concentrate on identity-discriminative signals on the query side, while performing holistic scene-level reasoning on the target side to accurately localize the queried identity. Together, these complementary capabilities form the basis of PS-ReID’s enhanced retrieval robustness across diverse and challenging scenarios.

Concretely, the query image $I^q$ and the target image $I^t$ are first processed by a shared vision encoder $f_{\Theta}$ (CLIP-ViT-L/14\cite{radford2021learning}) to extract high-level visual features. The features are then projected into the input embedding space of the language model $F_L$ ( LLaMA\cite{touvron2023llama}) through the role-specific adapters $\phi_R$ and $\phi_I$, achieving asymmetric multimodal alignment:
\begin{equation}
    x^q_{img}= \phi_R(f_{\Theta}(I^q))
    \label{eq:x_q_img}
\end{equation}
\begin{equation}
    x^t_{img}= \phi_I(f_{\Theta}(I^t))
    \label{eq:x_t_img}
\end{equation}

\noindent where $x^q_{img}$ and $x^t_{img}$ denote the image embeddings aligned to the language modality. 
In parallel, the textual instructions, which include the target description and user instruction, are tokenized into text embeddings via the word embedding layer $T$:
\begin{equation}
    x^q_{txt} = T(t^q) 
    \label{eq:tokenizer}
\end{equation}

Furthermore, PS-ReID introduces a structured multimodal prompt template $S$, where the placeholders $<image\_q>$ and $<image\_t>$ are replaced by $x^q_{img}$ and $x^t_{img}$. This design enforces explicit semantic separation between the query and target roles during reasoning, avoiding cross-role feature entanglement and improving identity matching accuracy. The LMM then performs autoregressive reasoning over the multimodal sequence:

\begin{equation}
    y = F_L(S(x^{q}_{img},x^{t}_{img}),x^q_{txt})
    \label{eq:tokenizer}
\end{equation}

Finally, a lightweight MLP text decoder transforms $y$ into the final output, with unified retrieval–segmentation control tokens coordinating cross-task supervision: [SEG] signals a successful match and triggers mask generation, [REJ] denotes absence, and [QRY] guides identity-specific reasoning for stronger ReID supervision.

\vspace{2mm}
\subsubsection{Segmentation Module}

While the segmentation module of PS-ReID is largely built upon the established SAM architecture \cite{kirillov2023segment}, its integration with the LMM introduces a seamless connection between reasoning and mask generation. Specifically, when the LMM predicts that the queried individual is present in the target image, it emits a dedicated control token [SEG], whose embedding $y_{seg}$ is projected into the prompt embedding space via a linear transformation $\phi$:
\begin{equation}
    h_{seg} = {\phi}(y_{seg})
    \label{eq:seg_embedding}
\end{equation}
where $h_{seg}$ serves as a semantic-aware prompt for the pixel decoder, enabling fine-grained localization amid cluttered conditions. Meanwhile, a frozen segmentation encoder $f_{S}$ extracts dense visual features from the target image $I^t$. These features, combined with the $h_{seg}$, are passed to the pixel decoder $f_d$ to generate the final binary mask $M$:

\begin{equation}
    M = f_d(h_{seg},f_{S}(I^t))
    \label{eq:tokenizer}
\end{equation}

Through this mechanism, the abstract decision of the LMM (via [SEG] token) is grounded into pixel-level localization, ensuring that the produced mask is not only spatially accurate but also identity-specific, aligned with the retrieval decision.

\subsection{Training Objectives}
To jointly optimize the reasoning, retrieval, and segmentation capabilities of our framework, we adopt a multi-task training strategy. The overall training objective consists of three components: a retrieval-guided ReID loss $\mathcal{L}_{\text{reid}}$ for aligning identity representations, a text generation loss $\mathcal{L}_{\text{text}}$ for supervising the output of the LLM, and a spatial-aware segmentation loss $\mathcal{L}_{\text{SA}}$ to guide the segmentation module in accurate mask prediction:

\begin{equation}
    \mathcal{L} = \mathcal{L}_{reid}+\mathcal{L}_{text}+\mathcal{L}_{SA}
    \label{eq:overall_loss}
\end{equation}

\noindent
We detail the design of each loss component in the following subsections.

\subsubsection{ReID Loss} 

Traditional ReID frameworks typically impose the loss on visual embeddings extracted from the backbone, keeping identity supervision detached from high-level reasoning and cross-modal interactions. In contrast, PS-ReID introduces a token-level supervision mechanism: instead of relying solely on backbone features, the ReID loss directly operates on the identity-aware control tokens produced by the LMM. We designate [QRY] as the token for the query identity, and [SEG] / [REJ] as the tokens for positive and negative samples, respectively. By integrating these control tokens into the ReID loss function, our approach aligns identity representation learning with the LMM’s decision space. This alignment enables the tokens to capture identity-discriminative information, which is directly utilized for generating identity-consistent masks in downstream tasks.

Formally, we randomly sample $P$ identities from the training set. For each identity, $K$ images are selected; one serves as the query (paired with its textual description) while the remaining images of the same identity are treated as positives, and images from other identities as negatives. Let $d_i^+$ and $d_i^-$ denote the cosine distances between the [QRY] token and its positive and negative sample tokens ([SEG] or [REJ]), respectively.
The ReID loss is defined as:

\begin{equation}
    \mathcal{L}_{reid} = \frac{1}{P}\lambda_{reid}\sum\limits_{i = 1}^P {\log (1 + {e^{\frac{{d_i^ +  - d_i^ - }}{\tau }}})} 
    \label{eq:reid_loss}
\end{equation}

\noindent
where $\tau$ controls the margin sharpness and $\lambda_{reid}$ is the weighting coefficient. To emphasize harder samples, $d_i^+$ is computed as a weighted sum:

\begin{equation}
    d_i^ + = \sum\limits_{j = 1}^K {w_{i,j}^+}{d_{i,j}^+} 
    \label{eq:d_i_+}
\end{equation}
\begin{equation}
    {w_{i,j}^+} = \frac{{\exp (d_{i,j}^ +  - \mathop {\max }\limits_m (d_{i,m}^ + ))}}{{\sum\limits_{k = 1}^K {\exp (d_{i,k}^ +  - \mathop {\max }\limits_m (d_{i,m}^ + )} )}}
    \label{eq:w_ij_+}
\end{equation}

The computation of $d_i^-$ follows the same formulation as $d_i^+$. This design emphasizes the harder positive and negative samples within each batch, encouraging the model to focus on more ambiguous instances and thereby enhancing its ability to learn fine-grained, discriminative features.

\subsubsection{Text Generation Loss}

The text generation loss adopts a standard autoregressive cross-entropy formulation, widely used in multimodal LLM training. Given a multimodal input consisting of the query image $I^q$, its textual description $t^q$, and the target image $I^t$, the LMM generates a textual response $r$ that is compared with the ground-truth annotation $a$:

\begin{equation}
    \mathcal{L}_{text} = \frac{1}{L} \sum_{j=1}^{L} -\log p(r_j \mid a_1, \ldots, a_{j-1})
    \label{eq:text_loss}
\end{equation}

This objective ensures coherent and accurate output while enforcing the correct production of control tokens ([SEG] or [REJ]). In the PS-ReID framework, it provides reliable supervision that anchors the reasoning process and supports both retrieval and segmentation tasks.

\subsubsection{Spatial-aware Segmentation Loss}

The spatial-aware segmentation loss is designed to supervise the segmentation module in producing accurate identity-consistent masks. It follows a standard multi-term formulation commonly adopted in segmentation tasks:

\begin{equation}
\begin{split}
\mathcal{L}_{SA} = & \ \lambda_{wbce} \mathcal{L}_{wbce} + \lambda_{dice} \mathcal{L}_{dice} \\
& + \lambda_{smooth} \mathcal{L}_{smooth} + \lambda_{ciou} \mathcal{L}_{ciou}
\end{split}
\end{equation}

\noindent
where $\mathcal{L}_{wbce}$ denotes the weighted binary cross-entropy loss, placing greater emphasis on foreground regions since pedestrian targets typically occupy a small portion of the image. $\mathcal{L}_{dice}$ measures overlap between the predicted mask and ground truth, encouraging precise boundary alignment. $\mathcal{L}_{smooth}$ introduces smooth L1 regularization, and $\mathcal{L}_{ciou}$ penalizes bounding-box misalignment through CIoU. Together, these components provide complementary supervision that drives the segmentation module to generate robust masks.

\section{Experiment}

\subsection{Experimental Setup}

\vspace{2mm}
\subsubsection{Data Split}
Our M\textsuperscript{2}ReID dataset is divided into four subsets: train, test, query, and gallery. The train set consists of 221,464 instances from 3,670 distinct identities, while the test set contains 91,695 instances from 1,224 identities. Importantly, there is no identity overlap between the training and test sets. The query set consists of 1,224 pedestrian slices, each paired with a corresponding textual description. These slices are randomly sampled from the test identities, with exactly one slice per identity. The gallery set is derived from the test set and includes three configurations: TestG50, TestG100 and TestG500, containing 50, 100 and 500 samples respectively. These variants are designed to evaluate the model's retrieval accuracy and segmentation performance under varying gallery scales.

\vspace{2mm}
\subsubsection{Evaluation Protocol}
We adopt the mean Averaged Precision (mAP), and the Cumulative Matching Characteristic (CMC) to evaluate the performance of person ReID. For person segmentation, we follow most previous works on referring segmentation\cite{kazemzadeh2014referitgame, mao2016generation} to adopt two metrics: gIoU and cIoU. Specifically, gIoU is defined as the average of the Intersection over Union (IoU) values across all images, while cIoU represents the cumulative IoU. Given that cIoU is sensitive to object size and tends to favor larger instances, we primarily report gIoU as the main evaluation metric for segmentation quality.

\vspace{2mm}
\subsubsection{Network Architecture}
We adopt the pre-trained LlaVA-llama2-13B or LLaVA-Lightning-7B-delta-v1-1\cite{liu2023visual} as the base Large Multimodal Models (LMMs) and apply LoRA~\cite{hu2022lora} with a rank of 8 for parameter-efficient fine-tuning. The vision encoder is a frozen CLIP-ViT-L/14-336 \cite{radford2021learning}.Both the R-T adapter and I-T adapter are composed of two linear layers and are trained from scratch. The segmentation module is implemented based on the SAM framework\cite{kirillov2023segment}.

\vspace{2mm}
\subsubsection{Implementation Details}
We conduct training using 4 NVIDIA A100 GPUs. The training process is divided into two stages. In the first stage, the adapters are trained from scratch. In the second stage, we fine-tune the pixel decoder and the large language model (LLM). The AdamW optimizer is employed, with learning rates set to 0.0003 and 0.00003 for the first and second stages, respectively. 
The weight of weighted binary cross-entropy (wBCE) loss, $\lambda_{wbce}$ set to 2.0, while all other loss weights are fixed at 1.0. During training, each batch consists of 3 identities, with 3 images sampled per identity.


\begin{table*}[ht]
\centering
\caption{Performance comparison between PS-ReID(ours) and existing methods on the M\textsuperscript{2}ReID dataset. \textbf{Bold} and \underline{underlined} indicate the best and second-best results, respectively. “ft” denotes fine-tuning on the M\textsuperscript{2}ReID training set.}
\vspace{2mm}
\label{table:PSReID_Results}
\begin{tabular}{ll|cccc|cccc|cccc}
\toprule

    \multicolumn{2}{c|}{\multirow{2}[4]{*}{Method}} & \multicolumn{4}{c|}{ TestG50 } & \multicolumn{4}{c|}{ TestG100 } & \multicolumn{4}{c}{ TestG500 } \\

    \cmidrule{3-14}
    \multicolumn{2}{c|}{} 
    & \multicolumn{1}{c}{gIoU} & \multicolumn{1}{c}{cIoU} & \multicolumn{1}{c}{top-1} & \multicolumn{1}{c|}{mAP}
    & \multicolumn{1}{c}{gIoU} & \multicolumn{1}{c}{cIoU} & \multicolumn{1}{c}{top-1} & \multicolumn{1}{c|}{mAP}
    & \multicolumn{1}{c}{gIoU} & \multicolumn{1}{c}{cIoU} & \multicolumn{1}{c}{top-1} & \multicolumn{1}{c}{mAP} \\
    
   \midrule

\multicolumn{2}{l|}{SEEM\cite{zou2023segment}} & 23.6 & \underline{37.2}  & - & - & 17.2 & 30.8  & - & - & 6.2 & 13.7  & - & -  \\
\multicolumn{2}{l|}{LISA-7B\cite{lai2024lisa}} & 12.6 & 12.4  & 10.0 & 16.7 & 9.1 & 9.3  & 6.9 & 8.3 & 5.0 & 5.1  & 3.4 & 5.7 \\
\multicolumn{2}{l|}{LISA-7B\cite{lai2024lisa}(ft)} & 17.4 & 17.6  & 12.5 & 19.4 & 12.8 & 12.9  & 10.3 & 13.0 & 7.8 & 7.6  & 6.5 & 8.9 \\
\multicolumn{2}{l|}{GSVA-7B\cite{xia2024gsva}} & 10.4 & 10.5  & 12.1 & 11.3 & 8.6 & 8.8  & 10.7 & 9.6 & 4.5 & 4.4  & 6.3 & 4.8 \\
\multicolumn{2}{l|}{GSVA-7B\cite{xia2024gsva}(ft)} & 18.0 & 18.6  & 24.1 & 21.7 & 15.4 & 15.7  & 20.8 & 17.4 & 11.6 & 12.0  & 16.4 & 14.4 \\
\rowcolor{gray!20}
\multicolumn{2}{l|}{PS-ReID-7B(ours)} & \underline{31.9} & 33.0  & \underline{40.3} & 29.3 & \underline{29.8} & \underline{31.2}  & \underline{36.3} & \underline{27.7} & \underline{20.0} & \underline{21.4}  & \underline{24.6} & \underline{22.1} \\
\midrule

\multicolumn{2}{l|}{LISA-13B\cite{lai2024lisa}} & 15.8 & 15.2  & 14.7 & 24.1 & 11.5 & 11.5  & 9.9 & 18.6 & 5.3 & 5.1  & 4.1 & 8.5 \\
\multicolumn{2}{l|}{LISA-13B\cite{lai2024lisa}(ft)} & 21.6 & 21.5  & 19.2 & 26.8 & 19.8 & 19.9 & 17.1 & 23.4 & 12.5 & 12.4 & 9.8 & 11.6 \\
\multicolumn{2}{l|}{GSVA-13B\cite{xia2024gsva}} & 14.6 & 14.8  & 16.6 & 13.0 & 9.5 & 9.8 & 12.8 & 10.2 & 5.6 & 5.5  & 7.8 & 5.3 \\
\multicolumn{2}{l|}{GSVA-13B\cite{xia2024gsva}(ft)} & 27.6 & 28.1  & 34.0 & \underline{30.8} & 23.7 & 24.0 & 28.4 & 26.9 & 18.2 & 18.5 & 20.7 & 17.1 \\
\rowcolor{gray!20}
\multicolumn{2}{l|}{PS-ReID-13B(ours)} & \textbf{57.6} & \textbf{52.1} & \textbf{81.2} & \textbf{66.3} & \textbf{54.7} & \textbf{50.8}  & \textbf{80.0} & \textbf{65.5} & \textbf{39.8} & \textbf{39.4}  & \textbf{72.1} & \textbf{60.0} \\

\bottomrule

\end{tabular}
\end{table*}


\subsection{Comparison to Existing Methods}

To our best knowledge, PS-ReID is the first framework to jointly address person ReID and segmentation task based on multimodal query inputs. To validate its effectiveness, we establish three baselines for comparative analysis.

\vspace{2mm}
\subsubsection{Baselines} 
\begin{itemize}
    \setlength{\itemsep}{0pt}
    \item SEEM\cite{zou2023segment}: A state-of-the-art image segmentation model that only supports mask generation. Therefore, its evaluation is limited to assessing mask quality on each benchmark.
    \item LISA\cite{lai2024lisa}: A single-target segmentation model that leverages SAM for mask prediction. Following the protocol in \cite{liu2023gres}, a predicted mask is considered empty if it contains fewer than 50 foreground pixels.
    \item GSVA\cite{xia2024gsva}: An extension of LISA that supports multiple reference masks simultaneously and explicitly handles unmatched targets by generating a special [REJ] token.
\end{itemize}

\vspace{2mm}
\subsubsection{Results} 
We compare our proposed method with the above baseline models to demonstrate its advantages in both re-identification and segmentation tasks. The results are shown in Table \ref{table:PSReID_Results}. To ensure a fair comparison, we report the zero-shot performance of LISA and GSVA, as well as their results after fine-tuning on the same training dataset. Since existing methods cannot simultaneously process multimodal queries and target images, we evaluate them using text-only queries. Although the image query of the target individual is omitted, resulting in a simpler setup compared to our method, the performance gap remains substantial. For instance, on the TestG500 set, PS-ReID-13B achieves 39.8\% gIoU, 39.4\% cIoU, 72.1\% top-1 accuracy, and 60.0\% mAP, respectively, significantly outperforming all baselines. This highlights the fundamental difference between our task and referring segmentation. These results highlight not only the essential differences between our task and conventional referring segmentation but also the critical role of multimodal queries in achieving accurate retrieval and segmentation.

In multimodal person ReID and segmentation, distinguishing individuals with subtle inter-class differences is particularly challenging. Relying solely on unimodal queries often leads to high false positive rates and unreliable identification. Consequently, although baseline methods perform well in referring segmentation, their effectiveness drops markedly in the multimodal ReID and segmentation setting. This degradation stems from the limited representational capacity of unimodal queries, which are insufficient to capture the fine-grained distinctions required for accurate identity matching.

Specifically, for SEEM\cite{zou2023segment}, which is a segmentation-only model without retrieval capability, we report only segmentation metrics (gIoU and cIoU). Although SEEM performs well in generic segmentation tasks, it lacks instance-level discrimination and frequently produces masks for non-existent individuals, particularly in the larger TestG500 set, leading to numerous false positives and reduced overall performance. 
For LISA\cite{lai2024lisa}, we apply its built-in mask score and area threshold to filter predictions. While effective in vision-language inference, it struggles to meet the combined demands of segmentation and retrieval under unimodal query inputs. 
For GSVA\cite{xia2024gsva}, the inclusion of the [REJ] token enhances its ability to filter out mismatched samples, resulting in improved performance over LISA after fine-tuning. However, its accuracy in both retrieval and segmentation still lags significantly behind our method, further demonstrating the necessity of multimodal queries for fine-grained identity understanding.

Overall, PS-ReID-13B consistently outperforms all baselines, while PS-ReID-7B achieves notable gains across most metrics. These results confirm the efficacy of our approach in retrieving target individuals and generating accurate masks in complex scenarios. We further analyze the model’s behavior through visualization and ablation studies in the following sections.

\begin{figure*}[ht]
\centering
\includegraphics[width=\linewidth]{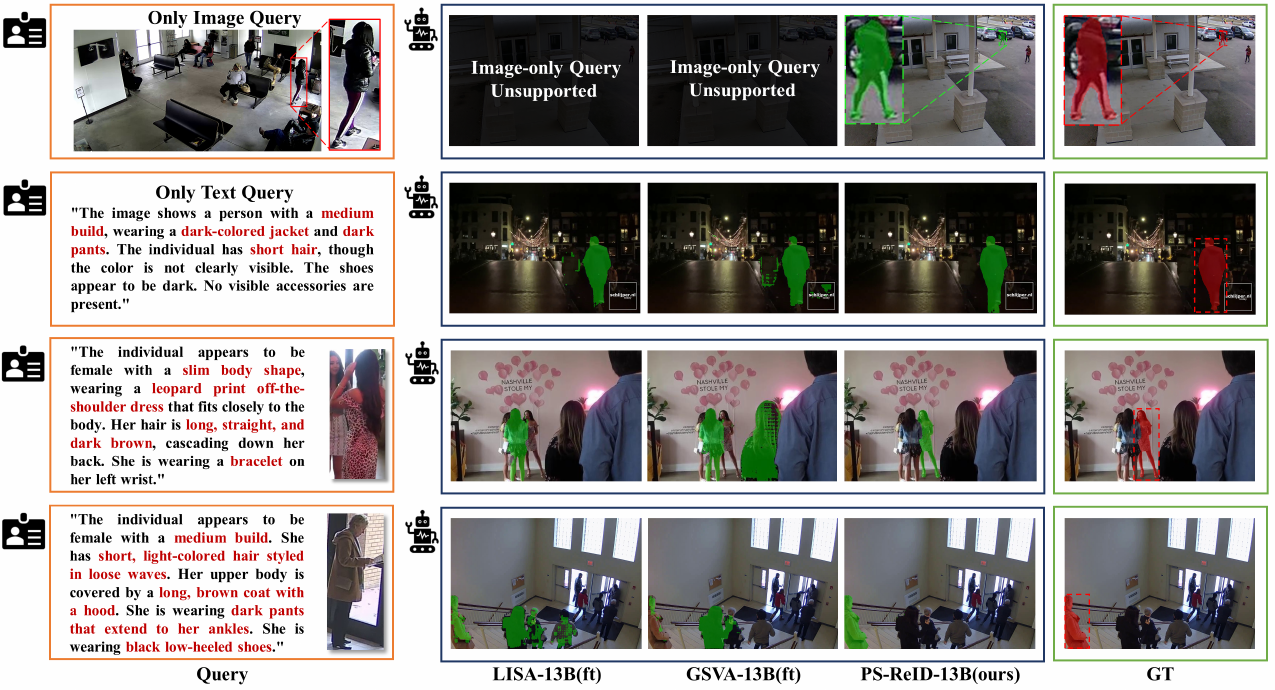}
\vspace{-0.25in}
\caption{Visual comparison among PS-ReID (ours) and existing related works. LISA and GSVA use text-only queries, whereas PS-ReID supports multimodal queries with both text and image inputs.}
\label{fig:compare}
\vspace{-0.15in}
\end{figure*}

\begin{figure}
\begin{center}
\includegraphics[width=1.0\linewidth]{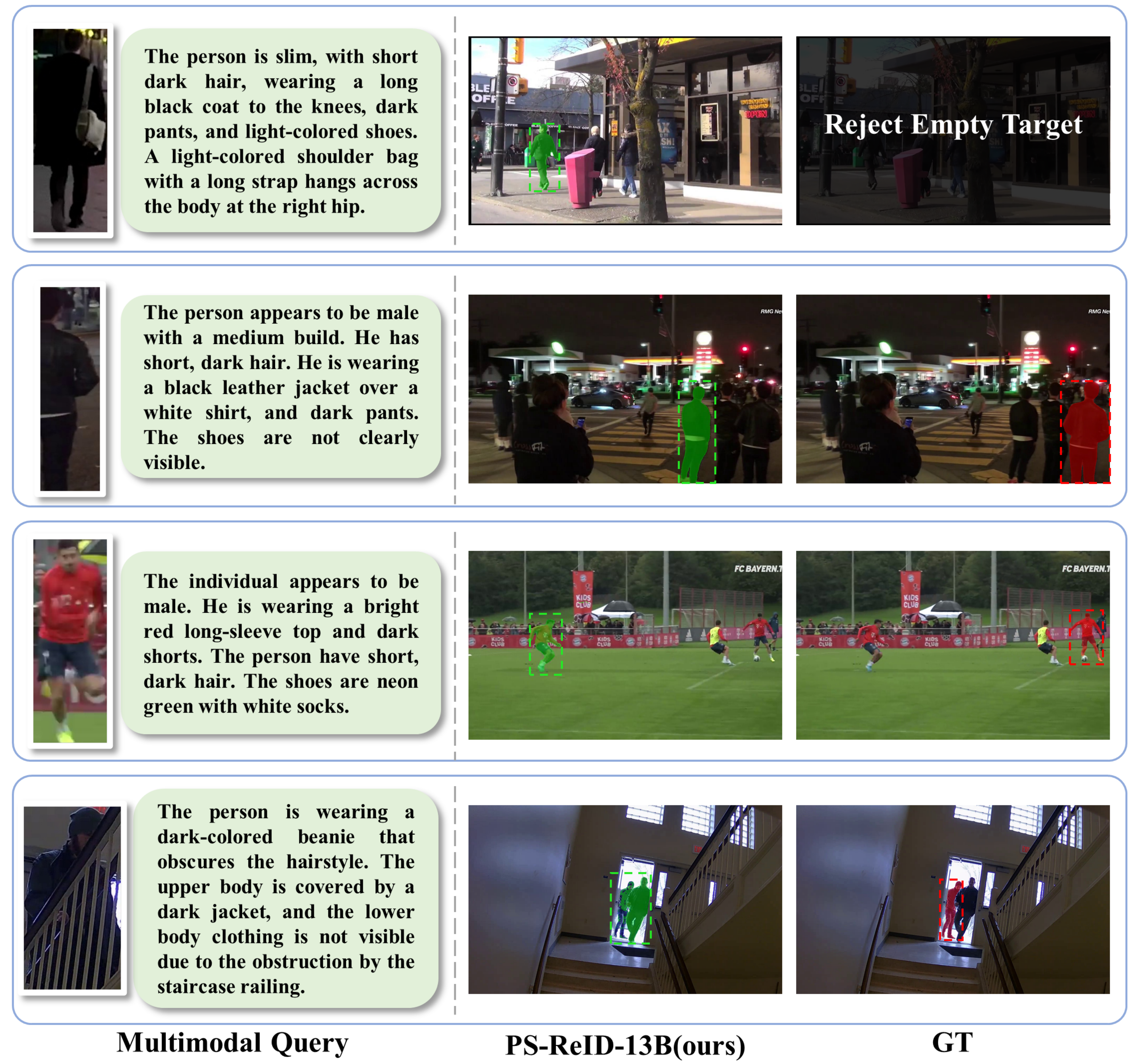}
\end{center}
\vspace{-0.15in}
   \caption{Visualization of PS-ReID failure cases. Typical errors occur when query and visual cues are insufficient, when multiple individuals share highly similar clothing and body shapes, or under severe occlusion, leading to mislocalization or inaccurate masks.}
\label{fig:FailureResult}\vspace{-0.15in}
\end{figure}

\vspace{2mm}
\subsubsection{Visualization Analysis}
Fig.~\ref{fig:compare} presents a qualitative comparison between our method (PS-ReID-13B) and two representative baselines, LISA-13B and GSVA-13B. In the image-only query setting, both baselines fail to produce outputs, whereas PS-ReID effectively interprets visual queries and generates accurate segmentation masks. In the text-only query setting, when the semantic cues are sufficiently informative to characterize the queried individual, all three methods exhibit comparable performance. However, in fine-grained identity discrimination scenarios involving multiple individuals, text-driven baselines often struggle to produce accurate masks when constrained to limited semantic cues. PS-ReID addresses this challenge by integrating visual evidence from query images with textual semantics, enabling the model to resolve ambiguities and generate masks that faithfully correspond to the queried identity.

The advantages of PS-ReID are more evident in challenging cases with occlusion and background clutter. Baseline methods often misidentify partially occluded individuals, while PS-ReID’s dual-path asymmetric encoding separates the roles of query and target: the query branch emphasizes identity-discriminative cues, whereas the target branch performs holistic scene reasoning. Combined with token-level supervision, the model enforces consistency between retrieval and segmentation, ensuring both are driven by the same identity signals. Consequently, PS-ReID delivers segmentation masks that combine spatial precision with identity consistency, demonstrating robustness across diverse and challenging conditions.

Despite its strong overall performance, PS-ReID still shows limitations in complex scenarios. As illustrated in Fig.~\ref{fig:FailureResult}, the model is prone to errors when queries are ambiguous, individuals share highly similar appearances, or severe occlusion occurs. These cases suggest that multimodal retrieval and segmentation remain challenging under ambiguity and limited information, motivating future research on stronger semantic disambiguation and context-aware modeling to further improve robustness.

\subsection{Ablation Study}

In this section, we conduct an extensive ablation study to reveal the contribution of each component. Unless otherwise specified, all results are reported using PS-ReID-13B on the M\textsuperscript{2}ReID dataset under the TestG100 gallery setting.

\vspace{2mm}
\subsubsection{Model Capacity}
To evaluate the effect of model scale, we compare our framework in two configurations: PS-ReID-7B and PS-ReID-13B. As shown in Table~\ref{table:PSReID_Results}, the 13B variant outperforms the 7B model across all metrics and gallery sizes, with particularly large margins under the TestG500 setting. The performance gains are not solely attributable to the increased number of parameters, but more critically stem from the 13B model's enhanced multimodal semantic reasoning capabilities. By effectively capturing the intricate relationships between visual content and textual descriptions, the 13B model demonstrates superior robustness in identity matching and segmentation accuracy, especially in challenging conditions such as occlusion and low lighting.

\vspace{2mm}
\subsubsection{Training Strategy}
We evaluate the impact of different training strategies: (1) training only the adapter, (2) jointly training the adapter and pixel decoder, and (3) further fine-tuning the LoRA-equipped LLM. As shown in Table~\ref{table:train_ablation}, training only the adapter enables basic retrieval and segmentation, but lacks sufficient semantic modeling to generate accurate masks. When the pixel decoder is jointly trained, the model shows significant improvements in capturing fine details and matching identities, leading to noticeable performance gains. Additionally, fine-tuning the LLM with LoRA results in steady improvements in both target retrieval and segmentation performance.

\begin{table}[ht]
\centering
\caption{Ablation study on the choice of training strategy.}
\vspace{2mm}
\label{table:train_ablation}
\resizebox{\linewidth}{!}{
    \begin{tabular}{ccc|cccc}
    \toprule
    
    \multicolumn{1}{c}{\multirow{2}[4]{*}{Adapter}} & \multicolumn{1}{c}{\multirow{2}[4]{*}{Pixel Decoder}} & 
    \multicolumn{1}{c|}{\multirow{2}[4]{*}{LLM(LoRA)}} &\multicolumn{4}{c}{ TestG100 } \\

    \cmidrule{4-7}

    \multicolumn{3}{c|}{} & \multicolumn{1}{c}{gIoU} &  \multicolumn{1}{c}{cIoU} & \multicolumn{1}{c}{top-1} & \multicolumn{1}{c}{mAP}  \\
 \midrule
 \checkmark & & & 37.1 & 37.6 & 58.1 & 42.0 \\
 \checkmark & \checkmark & & 50.2 & 46.7 & 75.3 & 56.8 \\
 \rowcolor{gray!20}
\checkmark & \checkmark & \checkmark & \textbf{54.7} & \textbf{50.8} & \textbf{80.0} & \textbf{65.5} \\
    \bottomrule
    \end{tabular}
}
\end{table}

\vspace{2mm}
\subsubsection{Query Modality}

To evaluate the contribution of multimodal queries, we compare three configurations: text-only, image-only and the combination of both modalities. As shown in Table~\ref{table:query_modality}, single modality queries perform notably worse. Text-only queries provide limited information, resulting in reduced performance, though they still outperform the baseline methods. In contrast, combining text and image queries significantly boosts performance, highlighting the complementary role of text in enriching the details captured by image queries. This underscores the substantial advantage of multimodal queries in improving the model's overall capability.

\begin{table}[ht]
\centering
\caption{Ablation study on query modality.}
\vspace{2mm}
\label{table:query_modality}
\begin{tabular}{p{2.5cm}|cccc}
\toprule
\multirow{2}{*}{\raisebox{-0.5ex}{Query Modality}} & \multicolumn{4}{c}{TestG100} \\
\cmidrule{2-5}
& gIoU & cIoU & top-1 & mAP \\
\midrule
Text only              & 30.0    & 31.3    & 45.9    & 30.9    \\
Image only             & 42.6   & 39.1   & 66.0   & 53.4   \\
\rowcolor{gray!20}
Text + Image           & \textbf{54.7} & \textbf{50.8} & \textbf{80.0} & \textbf{65.5} \\
\bottomrule
\end{tabular}
\end{table}

\vspace{2mm}
\subsubsection{Loss Functions}
We further investigate the impact of different loss components on model performance. While autoregressive loss, BCE loss, and Dice loss have been commonly utilized in previous works, our ablation study focuses on reid loss, smooth L1 loss, and CIoU loss. As shown in Table~\ref{table:loss_ablation}, comparing the first and fourth rows reveals that omitting the ReID loss leads to a significant performance drop across all metrics. This decline stems from the model's reduced ability to discriminate the target individual, making it challenging to accurately identify the target's presence in the scene, thereby increasing both false positives and false negatives. Introducing smooth L1 loss helps reduce prediction errors, improving both segmentation and target retrieval. Similarly, CIoU loss provides additional localization constraints, enhancing the model's ability to predict target position and size. These two losses complement each other, contributing to measurable improvements in both retrieval and segmentation performance.

\begin{table}[ht]
\centering
\caption{Ablation study on the choice of loss function.}
\vspace{2mm}
\label{table:loss_ablation}
\resizebox{\linewidth}{!}{
    \begin{tabular}{ccc|cccc}
    \toprule
    
    \multicolumn{1}{c}{\multirow{2}[4]{*}{ReID-L}} & \multicolumn{1}{c}{\multirow{2}[4]{*}{SmoothL1}} & 
    \multicolumn{1}{c|}{\multirow{2}[4]{*}{CIoU-L}} &\multicolumn{4}{c}{ TestG100 } \\

    \cmidrule{4-7}

    \multicolumn{3}{c|}{} & \multicolumn{1}{c}{gIoU} &  \multicolumn{1}{c}{cIoU} & \multicolumn{1}{c}{top-1} & \multicolumn{1}{c}{mAP}  \\
 \midrule
    & \checkmark & \checkmark & 45.4 & 43.5 & 68.7 & 50.1 \\
 \checkmark & & & 52.2 & 48.7 & 76.2 & 62.4 \\
 \checkmark & \checkmark & & 53.6 & 49.9 & 77.5 & 63.4 \\
 \rowcolor{gray!20}
\checkmark & \checkmark & \checkmark & \textbf{54.7} & \textbf{50.8} & \textbf{80.0} & \textbf{65.5} \\
    \bottomrule
    \end{tabular}
}
\end{table}

\section{Conclusion}

In this paper, we present PS-ReID, a novel multimodal framework that integrates image and text inputs to advance person ReID and segmentation. The proposed model overcomes the limitations of unimodal queries and further extends ReID to full-scene settings. The incorporation of segmentation enables precise representation of individuals when facing challenges such as occlusion. Our approach combines a dual-path asymmetric encoding that explicitly separates query and target semantics with token-level ReID supervision to enforce identity-consistent predictions. As a result, the model remains robust in scenarios involving background clutter and fine-grained identity discrimination.
To validate its effectiveness, we construct the M\textsuperscript{2}ReID dataset with high-quality segmentation masks and multimodal queries. Extensive experiments demonstrate that PS-ReID significantly outperforms existing unimodal models, achieving notable improvements in both ReID accuracy and segmentation quality. Overall, our findings highlight the crucial role of multimodal fusion in enhancing the accuracy and reliability of person retrieval and segmentation, offering a promising direction for future advancements in the field.

\bibliographystyle{IEEEtran}  
\bibliography{main}  

\vfill

\end{document}